\documentclass[10pt,twocolumn,letterpaper]{article}

\usepackage{cvpr}              

\usepackage[ruled,vlined,linesnumbered]{algorithm2e}

\usepackage{graphicx}
\usepackage{xcolor}
\usepackage{soul}
\usepackage{amsmath}
\usepackage{amssymb}
\usepackage{booktabs}
\usepackage{algpseudocode}
\usepackage{microtype}
\usepackage{float}
\usepackage{tabularx}
\usepackage{multicol}
\usepackage{tabularx}
\usepackage{lipsum}
\usepackage{nicefrac}

\DeclareMathOperator*{\argmin}{arg\,min}

\usepackage[pagebackref,breaklinks,colorlinks]{hyperref}

\usepackage[capitalize]{cleveref}
\crefname{section}{Sec.}{Secs.}
\Crefname{section}{Section}{Sections}
\Crefname{table}{Table}{Tables}
\crefname{table}{Tab.}{Tabs.}

\def\cvprPaperID{22} 
\def\confName{CVPR}
\def\confYear{2022}

\begin{document}

\title{
Overparameterization Improves StyleGAN Inversion
}

\author{Yohan Poirier-Ginter$^{\diamond}$, 
Alexandre Lessard$^{\bullet}$, 
Ryan Smith$^{\bullet}$, 
Jean-Fran\c{c}ois Lalonde$^{\diamond}$\\
$^\diamond$Universit\'e Laval, $^\bullet$Gearbox Software\\ 
\small{\texttt{\url{https://lvsn.github.io/OverparamStyleGAN/}}} 
} 

\twocolumn[{%
\renewcommand\twocolumn[1][]{#1}%

\maketitle
\vspace{-10mm}
\begin{center}
    \centering
    \captionsetup{type=figure}
    \includegraphics[width=\textwidth]{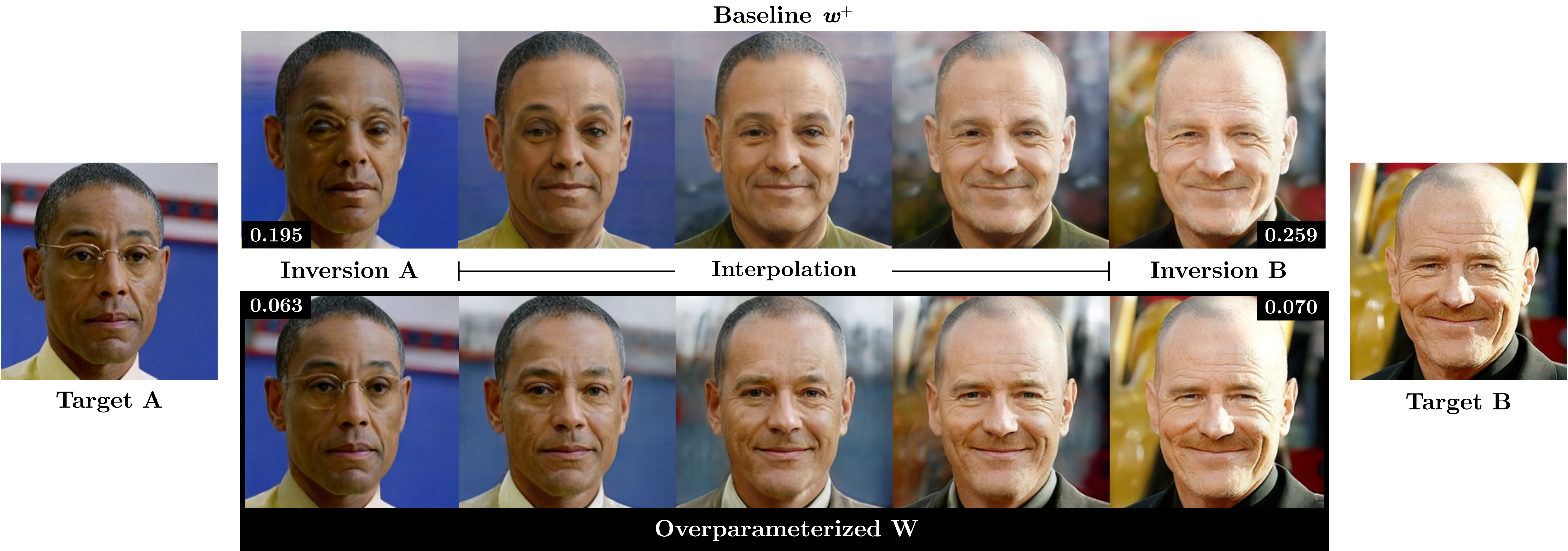}
    \captionof{figure}{
    Overparameterizing the StyleGAN latent space during training improves inversion quality. We obtain near-perfect reconstructions (``Inversion A'' and ``Inversion B'') of corresponding arbitrary real images (``Target A'' and ``Target B''), without having to sacrifice editability, as exemplified by interpolating between reconstructions (middle columns).}
\end{center}%
}]
\begin{abstract}
\vspace{-4pt}
Deep generative models like StyleGAN hold the promise of semantic image editing: modifying images by their content, rather than their pixel values. Unfortunately, working with arbitrary images requires inverting the StyleGAN generator, which has remained challenging so far. Existing inversion approaches obtain promising yet imperfect results, having to trade-off between reconstruction quality and downstream editability. To improve quality, these approaches must resort to various techniques that extend the model latent space after training. Taking a step back, we observe that these methods essentially all propose, in one way or another, to increase the number of free parameters. This suggests that inversion might be difficult because it is underconstrained. In this work, we address this directly and dramatically overparameterize the latent space, before training, with simple changes to the original StyleGAN architecture. Our overparameterization increases the available degrees of freedom, which in turn facilitates inversion. We show that this allows us to obtain near-perfect image reconstruction without the need for encoders nor for altering the latent space after training. Our approach also retains editability, which we demonstrate by realistically interpolating between images.
\end{abstract}
\vspace{-8pt}
\section{Introduction}

Deep image generation models~\cite{gan, biggan} ``fill the gaps'' between images of a set, making it possible to sample new, previously unseen images from the same domain. They also make possible the indirect manipulation of images via latent codes, rather than raw pixel values. In this space, StyleGAN~\cite{stylegan, stylegan2, stylegan2-ada, stylegan3} is a popular image generation model specifically because its choice of latent representation is auspicious for editing. Its latent code is mapped to style vectors which affect entire channels of feature maps, resulting in a spatially invariant parameterization. This learned mapping---from latent to style---is also thought to help with disentanglement~\cite{stylegan}, another necessity for image editing applications. In many ways, StyleGAN comes close to finally making possible the ambitious goal of semantic image editing. Unfortunately, there remains a missing piece: while edits work excellently for randomly sampled images, in its original formulation StyleGAN struggles with inversion, that is, reversing the generation process and finding a latent code that produces a real image of our choosing~\cite{inversion-survey}. Without this piece, StyleGAN is effectively only useful as a random image sampler.
\newpage

There have been many attempts at improving StyleGAN inversion. Here, there are two competing objectives at play: first, the reconstructed image must match the target as closely as possible (reconstruction quality); second, the obtained latent codes must be usable in downstream tasks (editability)~\cite{image2stylegan, interfacegan}. However, this second objective is not always met in practice: there exists multiple different latent codes that produce the same output image, and many of them produce unrealistic images when modified (\ie, they do not interpolate well with others)~\cite{image2stylegan, from-continu-to-edit, where-are-the-good-latents,gaussianized}.

Approaches to inversion can broadly be divided into two families: optimization and encoding.~\cite{inversion-survey}
Optimization finds latent codes by gradient descent, while encoding uses a trained network, called an encoder, to directly regress the latent code from an input image. It is generally considered that optimization reaches better reconstruction quality, while encoders provide better editability~\cite{where-are-the-good-latents,e4e}. In both cases, targeting StyleGAN's intermediate latent space $\mathcal{W}$ (the input of the synthesis network) works poorly: the obtained latent codes edit fairly well, but the reconstruction error is high~\cite{stylegan2}. This motivates different attempts at relaxing the target distribution. In particular, the extension to $\mathcal{W+}$ uses a different latent code for each layer. This change is a key component of most inversion pipelines~\cite{styleclip,stylegannada,stylerig,third-time,imageganmeet} as it improves reconstruction quality tremendously. But it comes at a price: because the latent space is altered after training, editability suffers. Hence, many approaches attempt to restrict this extension, trading-off between reconstruction quality and editability~\cite{e4e,pulse,where-are-the-good-latents,gaussianized}.

In this work, we propose a direct solution that avoids modifying the latent space after training: we dramatically overparameterize it beforehand. It turns out that large changes to the generator architecture are not required for this approach to train well, which we will show. We will then demonstrate how reconstruction quality is greatly improved following overparameterization. During inversion, our latent space is left as-is, which avoids the reconstruction quality/editability trade-off. Our optimization method only requires minimal regularization to achieve editability. In short, overparameterization of the latent space improves StyleGAN inversion.

\section{Previous work}
\label{previous-work}

\paragraph{Optimization} was the first approach to StyleGAN inversion~\cite{image2stylegan, image2stylegan++}. As compared to encoders, it does not require additional training and can dynamically alter its loss function. \textit{StyleGAN2}'s~\cite{stylegan2} inversion process obtained good editability but inadequate reconstruction quality, by targetting $\mathcal{W}$. \textit{Image2StyleGAN}~\cite{image2stylegan} first introduced the extension of $\mathcal{W}$ to $\mathcal{W+}$\footnote{In this work, we will usually refer to the elements of these spaces directly, i.e. vectors $\boldsymbol{w} \in \mathcal{W}$ and tuples of vectors $\boldsymbol{w}^+ \in \mathcal{W}^+$.}, greatly improving quality at the cost of editability. 
Some authors also optimize the noise inputs~\cite{stylegan2, image2stylegan++}, the feature maps~\cite{bdinvert, intermediate-optim, stylegan-of-all-trades, exploiting-spatial}, or even newly introduced transformations~\cite{transform-inverse, bdinvert}, usually in a second pass. Since then many papers propose regularization methods to improve editability following inversion~\cite{pulse, gaussianized, where-are-the-good-latents, indomain}. Finally, in \textit{Pivotal Tuning}~\cite{pivotal-tuning}, generator parameters are optimized along with latent codes, also in a second pass. In general, optimization obtains better reconstruction quality than encoders do, at least when targeting similar spaces. In this work, we focus on optimization, as it is the more flexible approach. We will also show that complex regularization is not required for retaining editability.

\paragraph{Encoders} directly learn a mapping between images and latent codes, and so inference takes milliseconds instead of minutes as with optimization. Typically, the encoder is trained after the fact, using a pretrained generator, although there have been attempts at training both jointly~\cite{ae-stylegan}. Most encoders employ a multi-level architecture, mirroring that of the generator, to target $\mathcal{W+}$~\cite{e4e,psp,ghfeat}, although some differ~\cite{face-identity}. Overall, similar editability concerns apply, and so some papers regularize the encoder to stay close to $\mathcal{W}$~\cite{e4e}. Further works improve quality through various techniques~\cite{restyle,a-simple-baseline,hyperstyle, collaborative}. In hybrid approaches, encoder initialization helps retain editability following an optimization-based refinement~\cite{indomain}. In general, it is thought that encoders obtain better editability than optimization, at the cost of flexibility. In this work, we argue that encoders are not strictly required for good editability.

\paragraph{Hypernetworks} predict the parameters of other networks~\cite{hypernetworks}; for StyleGAN specifically \textit{Hyperstyle}~\cite{hyperstyle} and~\cite{hyperinverter} use them to predict parameters offsets which fine-tune a generator, much like in \textit{Pivotal Tuning}~\cite{pivotal-tuning} except with an encoder. This approach is similar to our overparameterization, but, unlike us, they both target a pretrained StyleGAN generator while we retrain the generator following overparameterization. As such their latent space is different between training and inversion while ours remains as it was. Also similar to our work is \textit{AdaConv}~\cite{adaconv} for style transfer, but our modified style injection stays much closer to the original StyleGAN and they do not address inversion.\\

In summary, the vast majority of previous work aiming to improve reconstruction quality shares a point in common: it does so by increasing the number of targeted parameters in one way or another. Here, we take a more straightforward approach and instead, overparameterize the latent space directly. 
\newpage

\section{Method}
In this section, we show how the StyleGAN latent space can be overparameterized effectively. Altogether, these changes increase the latent dimensionality 512 times while avoiding redundant parameterization\footnote{We mean by this that overparameterization should gain additional degrees of freedom, i.e., naively increasing StyleGAN's latent size would not work as the projection layer would reduce its dimensionality back to a smaller size. }, excessive computational costs, or sweeping architectural changes.

\subsection{Overparameterized training}

\begin{figure}[h]
    \includegraphics[width=\linewidth]{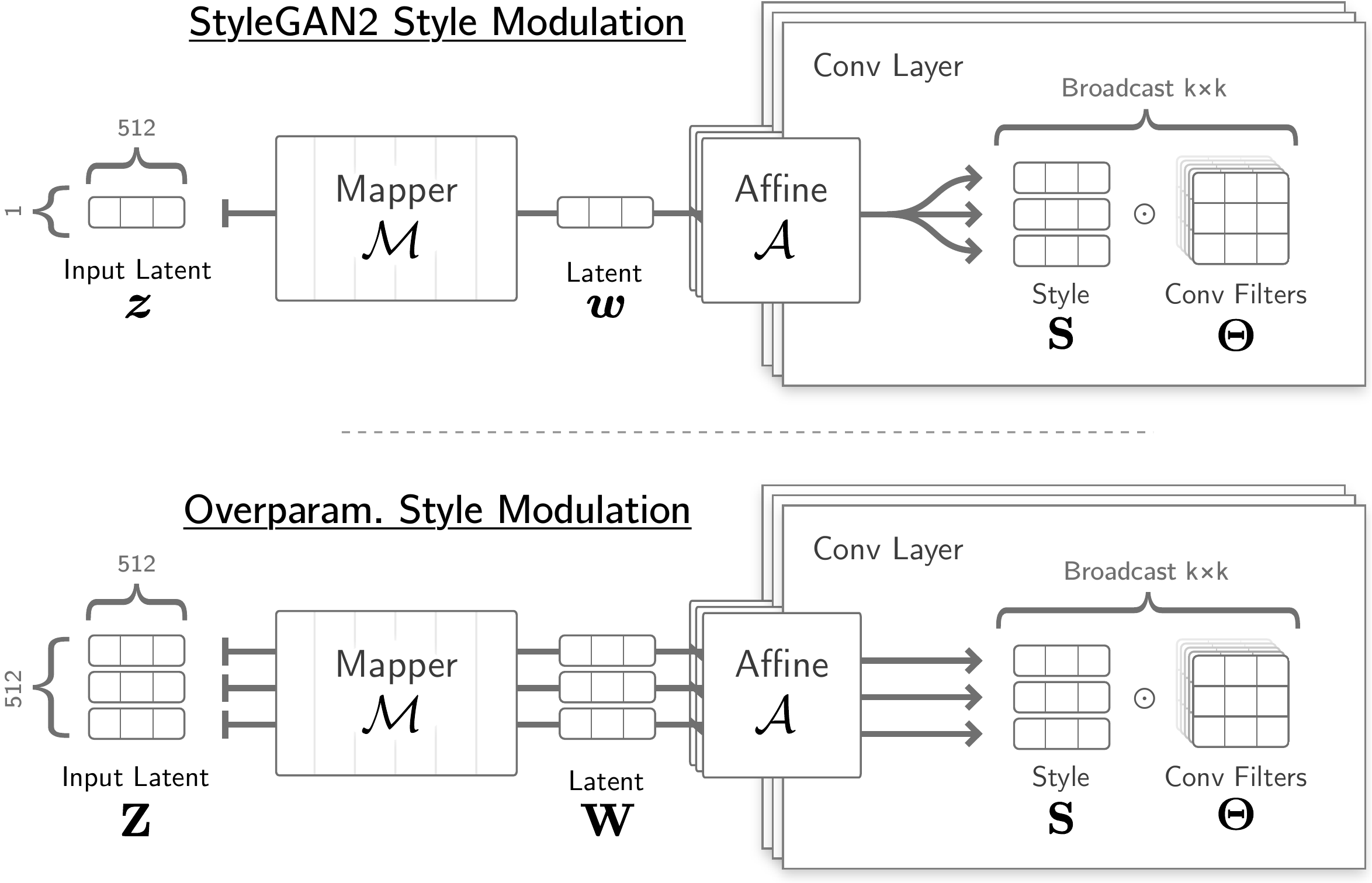}
    \caption{Modified style injection following overparameterization. StyleGAN2 first passes a single latent vector through mapping and affine projection layers. It then replicates the result row-wise before multiplication with the convolutional filters (top row). Instead, our approach instead maps and projects 512 vectors independently into a matrix of the same shape (bottom row).}\label{fig:method}
\end{figure}

\paragraph{StyleGAN2/3 style modulation}
For each layer $l$ of an $L$-layer network, StyleGAN2/3 style injection (fig.~\ref{fig:method}, top) modulates columns of the convolutional filter weights $\mathbf{\Theta}^{(l)} \in \mathbb{R}^{N_O^{(l)} \times N_I^{(l)} \times k^{(l)} \times k^{(l)}}$ 
with components of a style vector $\boldsymbol{s} \in \mathbb{R}^{N_I^{(l)}}$, replicating across other dimensions:
\begin{equation}
\mathbf{\Theta'}^{(l)}_{:, j, :, :} = \mathbf{\Theta}^{(l)}_{:, j, :, :} \cdot \boldsymbol{s}^{(l)}_{j} \,.
\label{baseline-modulation}
\end{equation}
Here, $N_I^{(l)}$,  $N_O^{(l)}$, and $k^{(l)}$ refer to the number of input channels, output channels, and the kernel size, respectively, of layer $l$. The style vectors $\boldsymbol{s}^{(l)}=\mathcal{A}^{(l)}(\mathcal{M}(\boldsymbol{z}))$ are obtained by passing $\boldsymbol{z} \in \mathbb{R}^{512} \sim \mathcal{N}(0, I)$ through a shared mapping network $\mathcal{M} : \mathbb{R}^{512} \rightarrow \mathbb{R}^{512}$ followed by layer-specific affine projections $\mathcal{A}^{(l)} : \mathbb{R}^{512} \rightarrow \mathbb{R}^{N_I^{(l)}}$. In particular, the same style vector is used for all rows of a layer convolution filter weights.

\paragraph{Overparameterized style modulation}
\label{method-overparam-modulation} 

In contrast, our approach samples a \emph{different style for each row}, as depicted in fig.~\ref{fig:method} (bottom). Grouping 512 different latent vectors into the rows of $\mathbf{Z} \in \mathbb{R}^{512 \times 512}$, we map and project each of them independently into a matrix $\mathbf{S}^{(l)} \in \mathbb{R}^{N_o^{(l)} \times N_i^{(l)}}$, where $\mathbf{S}^{(l)}_i = \mathcal{A}^{(l)}(\mathcal{M}(\mathbf{Z}_i))$\footnote{When $N_O^{(l)} < 512$, we take the easy way out and drop rows.}. Style modulation is unchanged except that it now also varies row-by-row:
\begin{equation}
\mathbf{\Theta'}^{(l)}_{i, j, :, :} = \mathbf{\Theta}^{(l)}_{i, j, :, :} \cdot \mathbf{S}_{i,j}^{(l)} \,.
\label{overparam-modulation}
\end{equation}

For training to be effective, we found it essential for latent codes to share some correlation. From a shared vector $\boldsymbol{\tilde{z}} \sim \mathcal{N}(0, I)$ and 512 i.i.d. vectors $\boldsymbol{\tilde{z}}^{(i)} \sim \mathcal{N}(0, I)$, we set $\mathbf{Z}_i = (\boldsymbol{\tilde{z}}^{(i)} + \boldsymbol{\tilde{z}}) / \sqrt{2}$. With this approach, each latent $\mathbf{Z}_i$ is still a unit Gaussian. As such, the distribution of the mapper's output $\mathcal{W}$ is unique; each row of $\mathbf{Z}$ is identically distributed, but non-independent. We will show in sec.~\ref{ganspace} that this allows compatibility with existing editing approaches. As an added benefit, calls to the mapping and projection layers can vectorized, reducing computational costs (sec.~\ref{comp-eff}). 

\subsection{Overparameterized inversion}

\begin{figure}[h]
    \centering
    \includegraphics[width=1\linewidth]{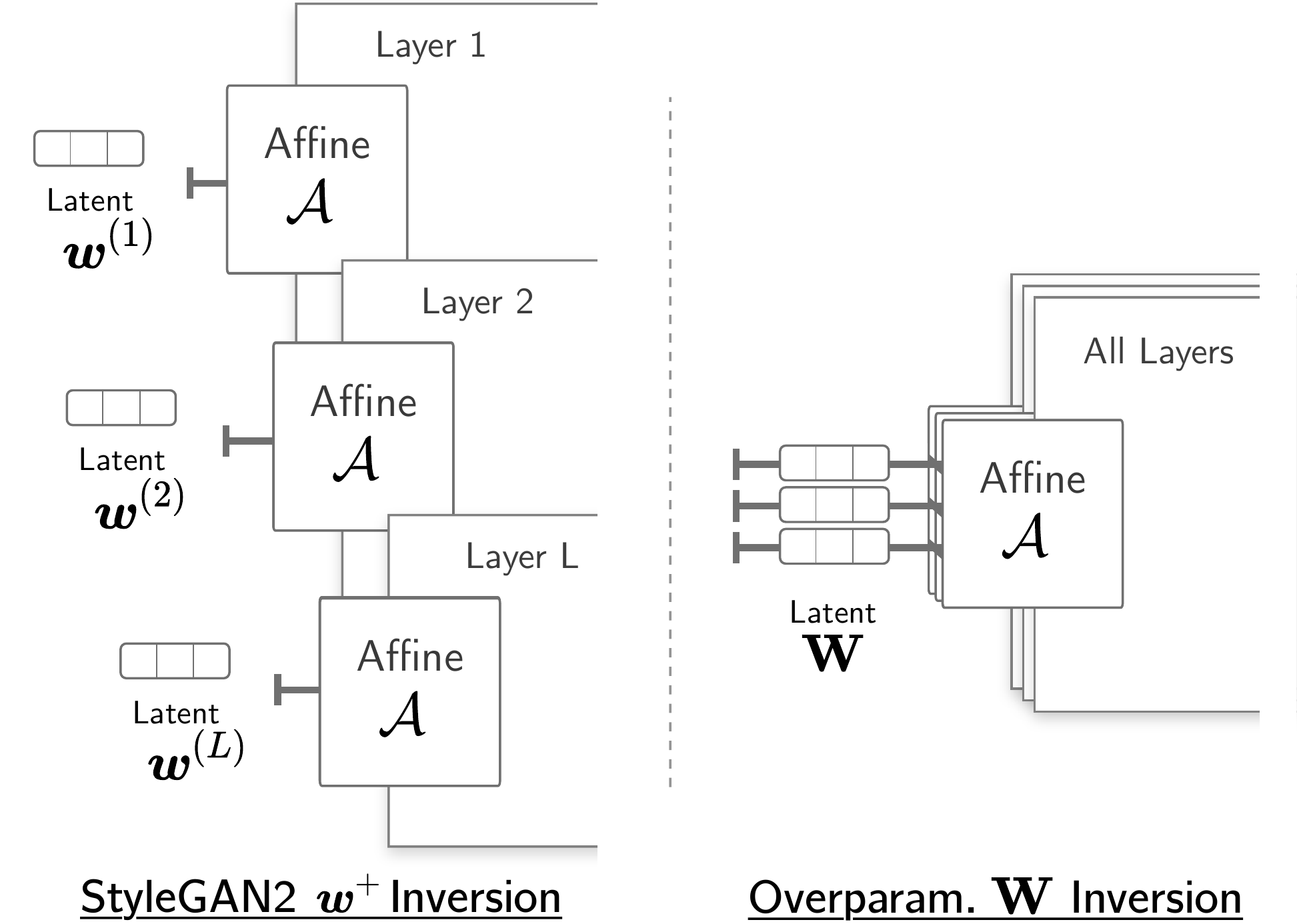}
    \caption{
    Targeted latent space during inversion. To obtain reconstructions of good quality, (left) StyleGAN must modify the latent space after training, and target different latent codes for each layer. (right) In our overparameterized network, we simply target the matrix $\mathbf{W}$, and inject it just like during training.}\label{fig:method-inversion}
\end{figure}

\paragraph{StyleGAN inversion} As described in sec.~\ref{previous-work}, naive StyleGAN inversion targets intermediate latent codes $\boldsymbol{w} \in \mathcal{W}$, ignoring the mapping network. It attempts to reconstruct a latent code $\boldsymbol{w}$ which generates a target image $\boldsymbol{y}$ by iteratively solving for 
\begin{equation}
\argmin_{\boldsymbol{w}} \mathcal{L}(\mathcal{S}(\boldsymbol{w}, ..., \boldsymbol{w}), \boldsymbol{y}) \,,    
\end{equation}
where $\mathcal{S}$ is the synthesis network (a function which accepts one latent code per layer and outputs an image), $\mathcal{L}$ is a loss function (typically comprised of a perceptual distance~\cite{lpips} and sometimes other terms) and $\boldsymbol{w}$ is initialized to $\mu_{\mathcal{W}} = \mathbb{E}_{\boldsymbol{z}}[\mathcal{M}(\boldsymbol{z})]$, the mean in $\mathcal{W}$. Unfortunately, this approach performs poorly and fails to match the target. Empirically, it was found that optimizing a separate latent code for each layer, \ie,
\begin{equation}
\argmin_{\boldsymbol{w}^+=(\boldsymbol{w}^{(1)}, ..., \boldsymbol{w}^{(L)})} \mathcal{L}(\mathcal{S}(\boldsymbol{w}^{(1)}, ..., \boldsymbol{w}^{(L)}), \boldsymbol{y}) \,,
\end{equation}
obtains much better reconstruction quality~\cite{image2stylegan} (see fig.~\ref{fig:method-inversion}, left). However, this change also modifies the latent space after training, which in turn damages editability\footnote{Avoiding this by training with i.i.d. vectors for each layer leads to a degradation of performance, motivating other solutions~\cite{w++}. At the same time, it is desirable to even further increase the number of parameters, justifying our final approach.}.

\paragraph{Overparameterized inversion} In our case, we sample a latent matrix $\mathbf{W}$ (fig.~\ref{fig:method-inversion}, right), where the rows are correlated samples from $\mathcal{W}$, as described previously. An analogous extension to $\mathbf{W}^+ = (\mathbf{W}^{(1)}, ..., \mathbf{W}^{(L)})$ is not required, as $\mathbf{W}$ already contains enough parameters for a fit of good quality. Initializing each row of $\mathbf{W}$ to $\mu_{\mathcal{W}}$, we can directly solve for
\begin{equation}
\min_{\mathbf{W}} \mathcal{L}(\mathcal{S}(\mathbf{W}, ..., \mathbf{W}), \boldsymbol{y}) \,,
\end{equation}
where $\mathbf{W}$ is shared between layers as during training. The matrix $\mathbf{W}$ contains $\nicefrac{512}{L}$ more parameters than $\boldsymbol{w}^+$, which greatly improves reconstruction. We will demonstrate this in the following section, while also showing that editabiliy is retained.
\section{Experiments}

This section will first show how generation quality is mostly retained even with our (admittedly quite drastic) overparameterization. Then, we will show how inversion is improved: reconstruction of much better quality is obtained while editability is preserved.

\subsection{Generation quality}

We train \textit{stylegan2-ada-pytorch}\cite{stylegan2-ada} at a resolution of $256\times256$ on the FFHQ dataset~\cite{stylegan}. Starting from the default configuration, we modernize the architecture following StyleGAN3~\cite{stylegan3}: we omit the noise inputs (they are a hindrance in image editing applications) but retain the perceptual path length (PPL) regularizer, which improves editability and inversion quality~\cite{stylegan2}. We also keep the style mixing regularizer, as style mixing has use in many downstream tasks. Finally, we reduce the number of mapping layers from 8 to 2, which was shown to improve FID~\cite{stylegan3}. We do not specialize hyperparameters for our method, nor do we specialize them for the dataset\footnote{We performed all trainings using the default configuration, rather than the configuration specialized for FFHQ.}. The same configuration is used for training both the baseline and our method.

\begin{figure}[t]
    \vspace{-2.00mm} 
    \includegraphics[width=\linewidth]{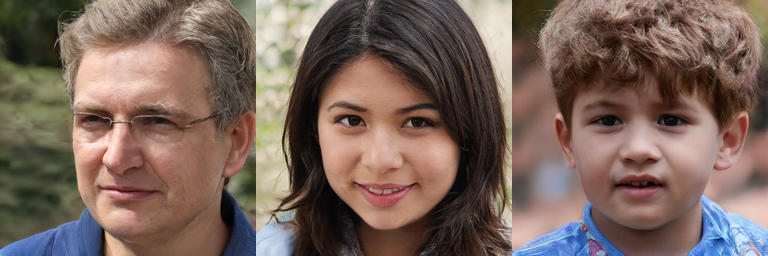}
    \caption{Selected generated images sampled from our overparameterized model, with a resolution of $256\times256$. Each image has $512\times512$ latent parameters, more than the number of pixels.}\label{fig:generation_quality}
\end{figure}

We measure the generation quality using the standard Frechet Inception Distance~\cite{fid} (FID) for generation quality and the Perceptual Path Length~\cite{stylegan2} (PPL) for disentanglement. After completing training, we retain the checkpoint with the lowest FID, comparing 50k generated images to 50k images of the training set. Note that comparing the PPL between two approaches with different latent space dimensionality is valid, as the PPL is an image-space metric despite measuring sensitivity to latent changes.

\begin{table}[h]
\vspace{-2.00mm} 
\small
\centering
\begin{tabularx}{\linewidth}{Xcc}
\toprule

& \textbf{FID}$_\downarrow$ (Realism) &  \textbf{PPL}$_\downarrow$ (Disent.) \\
\midrule
Baseline & 3.88 & 196.27 \\
Overparam & 5.05 & 203.67\\
\bottomrule
\end{tabularx}
\caption{Generation quality on the FFHQ dataset. FID (quality) at the lowest point during training and corresponding PPL (disentanglement), for both methods.}
\label{table:generation_quality_metrics}
\end{table}

Table~\ref{table:generation_quality_metrics} shows that the FID score does quite not match the baseline; there is some degradation in generation quality. Nevertheless, most images are still very realistic, as shown in fig.~\ref{fig:generation_quality}. Table~\ref{table:generation_quality_metrics} also shows that disentanglement is essentially unchanged, which is important for downstream tasks.

\subsection{Reconstruction quality}\label{recon-quality}

\begin{figure}[t]
    \includegraphics[width=\linewidth]{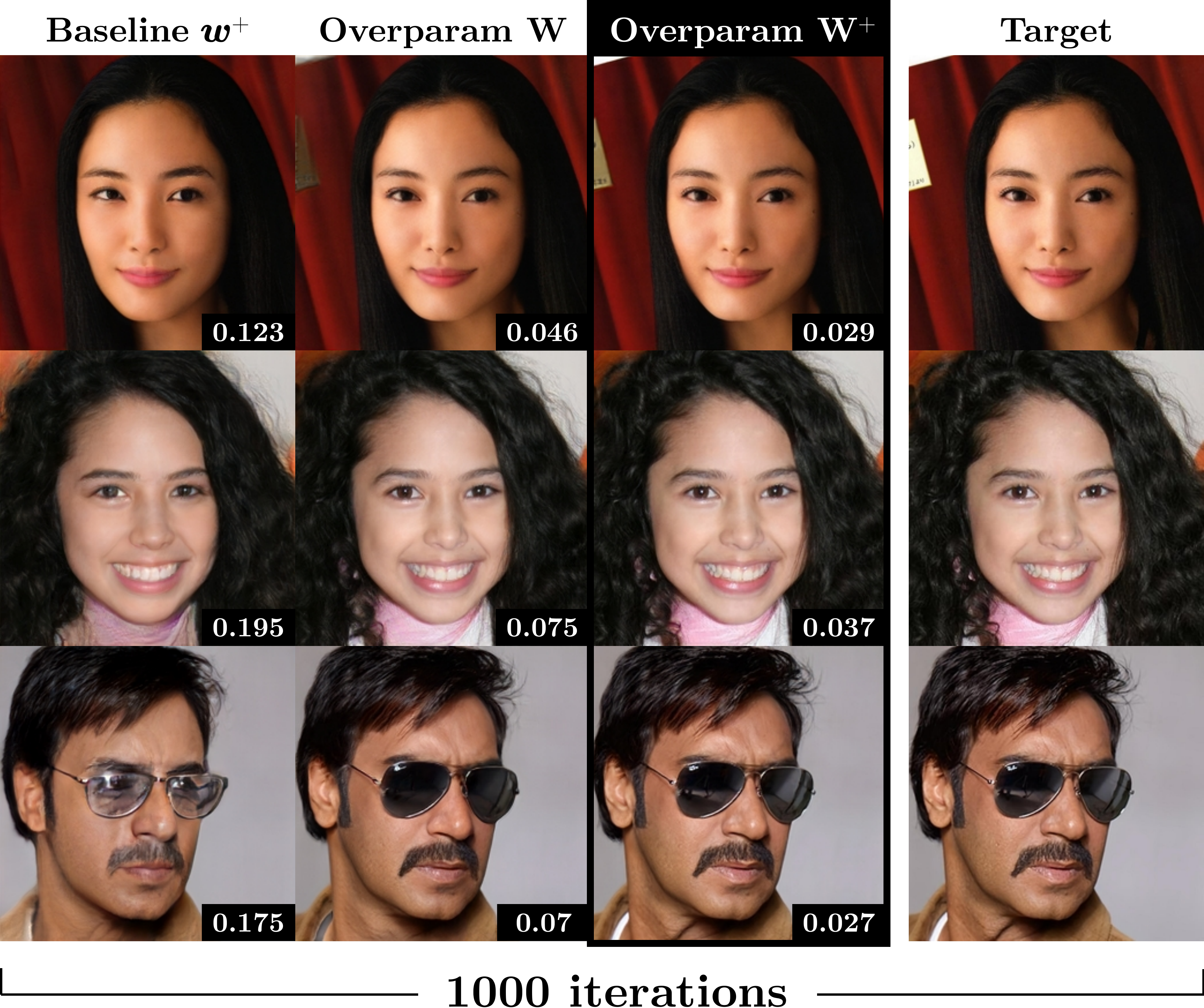}
    \caption{Reconstruction quality after inversion for selected examples. We compare inversion targeting $\boldsymbol{w}+$ for the baseline (first column) to inversion targeting $\mathbf{W}$ and its extension $\mathbf{W}^+$ for our method (second and third row, respectively). Inlays report the LPIPS metric between the reconstruction and target.} \label{fig:recon_quality}
\end{figure}

\begin{figure}[h]
    \centering
    \includegraphics[width=1\linewidth]{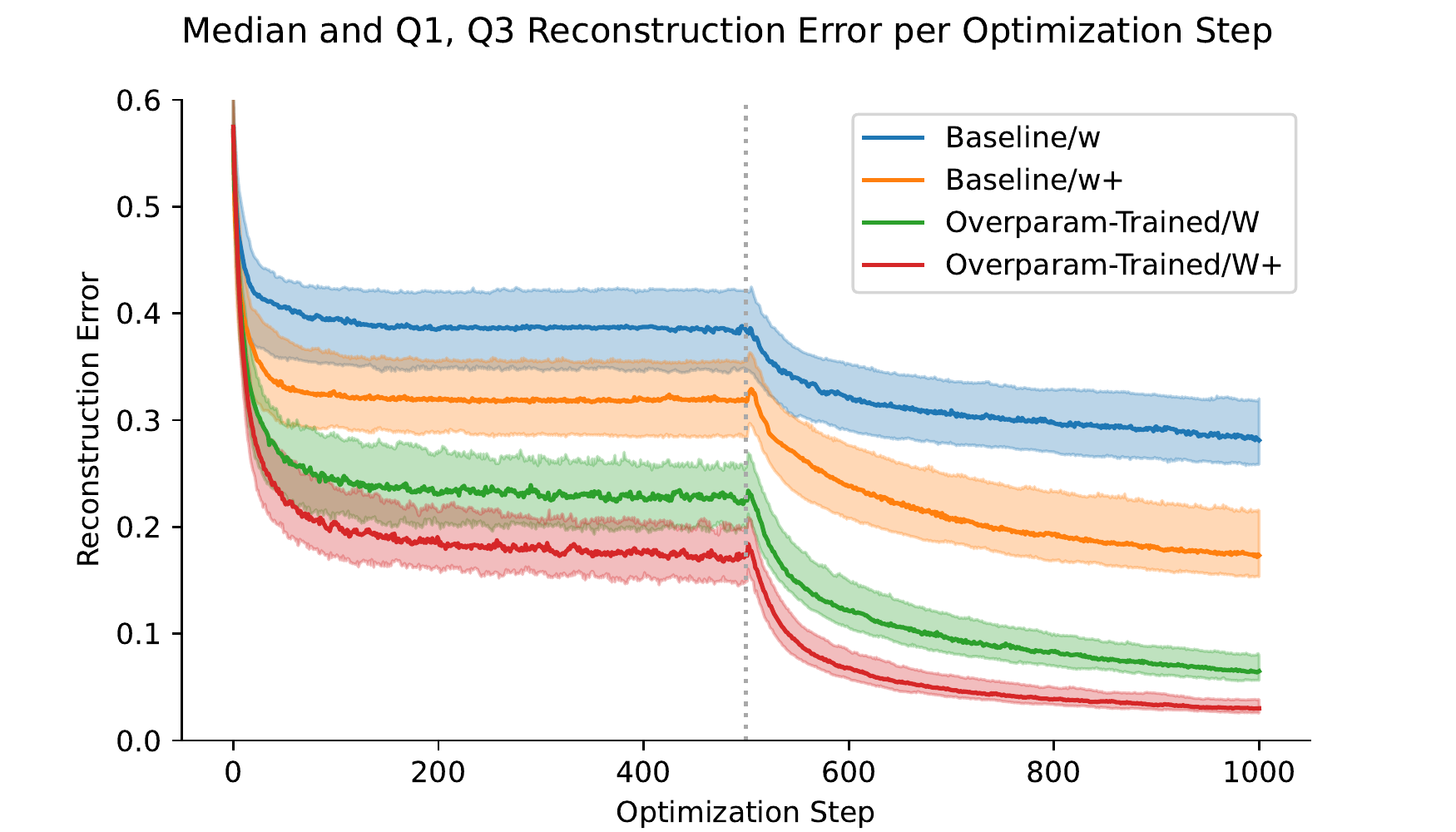}
    \caption{Median reconstruction quality (LPIPS) and quantiles measured over 100 CelebA-HQ images, when targeting different spaces. The dotted line shows when regularization was disabled.}\label{fig:losses_plot}
\end{figure}

We measure reconstruction quality using optimization, on 100 images of the CelebA-HQ dataset\footnote{GANs metrics are typically reported after training on entire datasets (FFHQ in our case). As such, we evaluate on a separate, similar dataset (CelebA-HQ).}~\cite{progan, celeba}. We optimize for 1000 steps with the Adam optimizer, using settings $\eta=0.05, \beta_1=0.9, \beta_2=0.999$, and the LPIPS loss from StyleGAN2. We compare the baseline $\boldsymbol{w}$ and its extended version $\boldsymbol{w}^+$, to our overparameterized $\mathbf{W}$ and its extended version $\mathbf{W}^+$. We regularize with a $0.9$ truncation~\cite{stylegan} towards $\mu_{\mathcal{W}}$ before each optimization step, disabled halfway.\footnote{Hybrid approaches edit well following encoder initialization. We essentially replace encoder initialization with regularized initialization.}
As table \ref{table:reconstruction_quality_metrics} shows, overparameterization improves reconstruction quality by large amounts, even without extension to layer-specific codes. That being said, such an extension improves quality even more. Fig.~\ref{fig:recon_quality} shows examples of our method recovering all important details of the target image, even for harder instances. In addition, fig.~\ref{fig:losses_plot} shows that our method converges faster and obtains great reconstructions with high reliably.

\begin{table}[ht]
\small
\centering
\begin{tabularx}{\linewidth}{Xcc}
\toprule
&\textbf{LPIPS}$_\downarrow$ (Distort.)  & \textbf{FID}$_\downarrow$ (Realism)\\
\midrule
Baseline $\boldsymbol{w}^+$ & 0.185 & 77.16\\ 
Overparam $\mathbf{W}$ & 0.070 & 73.52\\
Overparam $\mathbf{W}^+$ & 0.033 & 74.19\\
\bottomrule
\end{tabularx}
\caption{Reconstruction quality on 100 images of the CelebA-HQ dataset. We compare the baseline (first row) to our overparameterization, with and without extension to layer-specific codes (third and second rows, respectively).}
\label{table:reconstruction_quality_metrics}
\vspace{-4mm}
\end{table}

\subsection{Real image interpolation}

Because image editing can be achieved using linear combinations of latent vectors~\cite{interfacegan}, as is the case with interpolation, we use interpolation quality on inverted images as a proxy measure for editability~\cite{e4e}. First, we generate interpolations between all pairs of our inverted images from sec.~\ref{recon-quality}, yielding 4,950 images. We then report the FID of the midpoint image distribution, compared to the distribution of the full CelebA dataset, and the perceptual path length over the entire interpolation sequence\footnote{More precisely, we divide each interpolation path into 5 segments and sum the total (perceptual) variation, averaging over all images.}. 

\begin{figure}[h]
    \includegraphics[width=\linewidth]{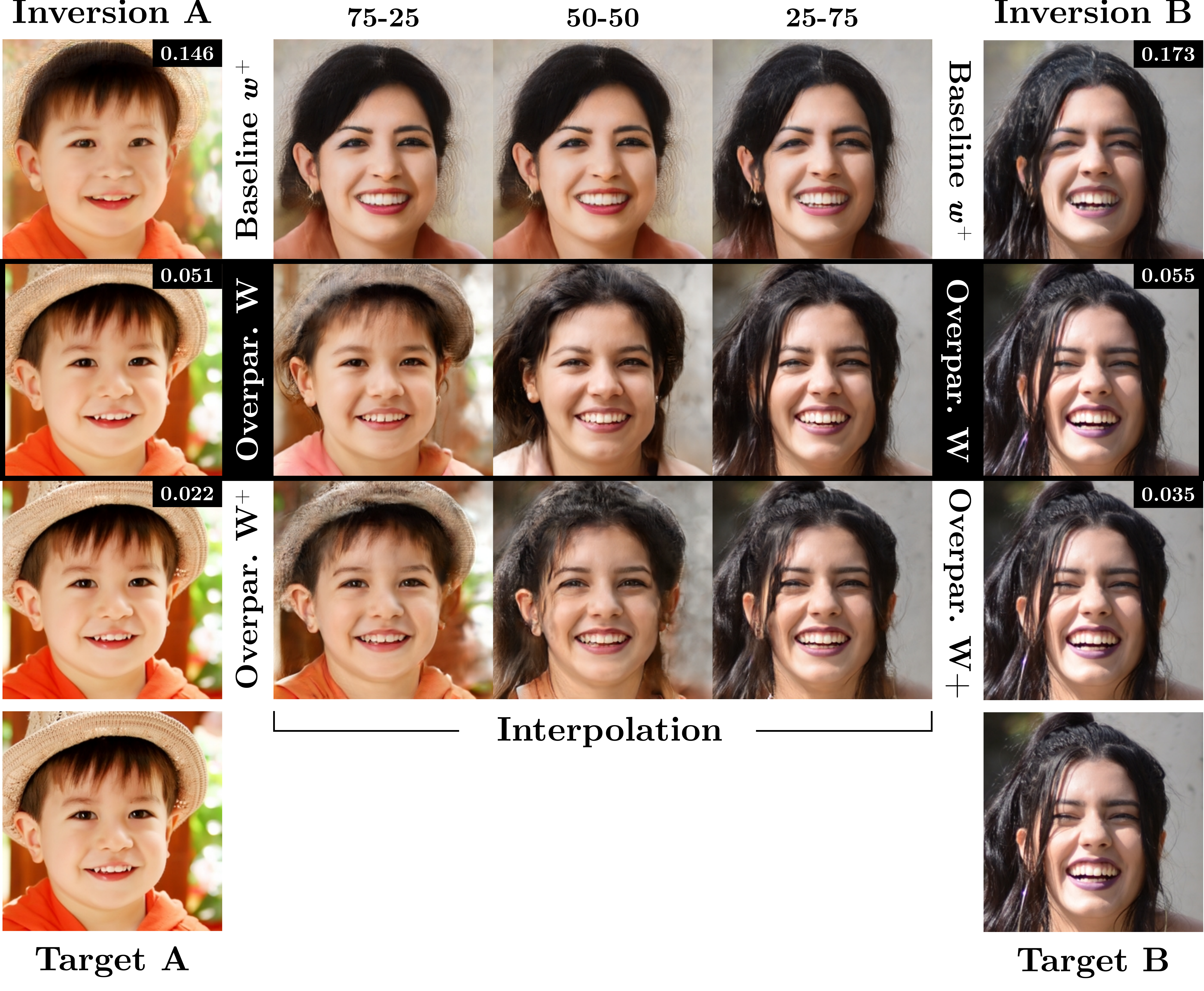}
    \caption{Interpolation quality for selected examples. We compare inversion targeting $\boldsymbol{w}+$ for the baseline (first row) to inversion targeting $\mathbf{W}$ for our method (second row). Inlays report the LPIPS metric between the reconstruction and target. Extension to $\mathbf{W}^+$ produces interpolations with more artifacts (third row).} \label{fig:interpolation_quality}
\end{figure}

Table \ref{table:interpolation_quality} shows that overparameterization does not compromise editability significantly despite reaching much better reconstruction quality, as shown previously. Unsurprisingly, extending the latent space after training gives worse editability. Our inversion better matches the targets, which helps the midpoint image retain characteristics of both endpoints (fig.~\ref{fig:interpolation_quality}). However, this does not appear to be reflected the PPL, which remains relatively constant, albeit slightly better for the baseline.  Refer to the supplemental for additional qualitative results on image interpolation. 

\begin{table}[H]
\small
\centering
\begin{tabularx}{\linewidth}{Xcc}
\toprule
&\textbf{FID}$_\downarrow$ (Realism) & \textbf{PPL}$_\downarrow$ (Disent.) \\
\midrule
Baseline $\boldsymbol{w}^+$ & 52.97 & 0.175\\
Overparam $\mathbf{W}$ & 53.51 & 0.190\\
Overparam $\mathbf{W}^+$ & 57.60 & 0.188\\
\bottomrule
\end{tabularx}
\caption{Interpolation quality on all pairs from 100 images of the CelebA-HQ dataset. We compare the baseline (first row) to our overparameterization, with and without extension to layer-specific codes (second and third rows, respectively).}
\label{table:interpolation_quality}
\end{table}

\section{Compatibility with downstream tasks}

In this section, we briefly demonstrate the compatibility of
our modified latent space in existing downstream tasks; this
section intends to show that previous methods still work.

\subsection{Generator-driven upsampling (PULSE)}

In PULSE~\cite{pulse}, optimization searches for the closest match after the generated image is also downsampled. Such a problem is underconstrained and requires more regularization; we use the same optimization process but keep the truncation regularizer throughout. We show preliminary results with $16\times$ nearest-neighbour downsampling.

\begin{figure}[h]
    \includegraphics[width=\linewidth]{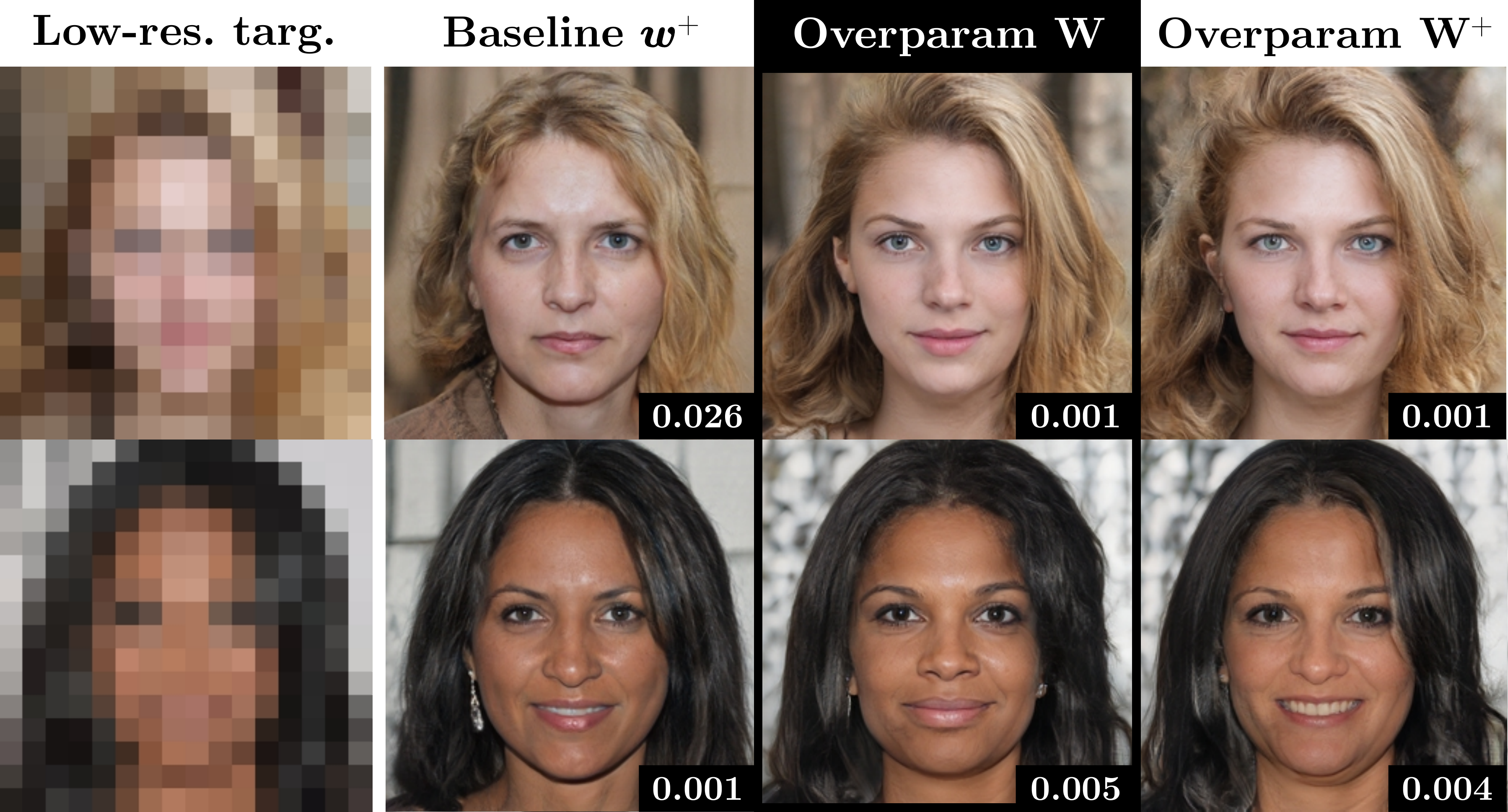}
    \caption{Generator-driven image upsampling. The optimization searches for an image that matches the low-resolution target on the left, once also downsampled. Inlays give the LPIPS between the downsampled targets and the downsampled reconstructions.}\label{fig:applications}
\end{figure}

Fig.~\ref{fig:applications} show that our method obtains great results, which match the low-resolution target more closely than the baseline. While the extension of our method to layer-specific codes still obtains an even better fit, the faces seem slightly less realistic; please also refer to fig.~\ref{fig:supplemental-editing} for additional results, which also shows the original (unknown) full-resolution target. As described previously in sec.~\ref{recon-quality}, our method also converges to adequate results much faster.

\subsection{Unsupervised semantic editing (GANSpace)}
\label{ganspace}
\begin{figure}[h]
    \includegraphics[width=\linewidth]{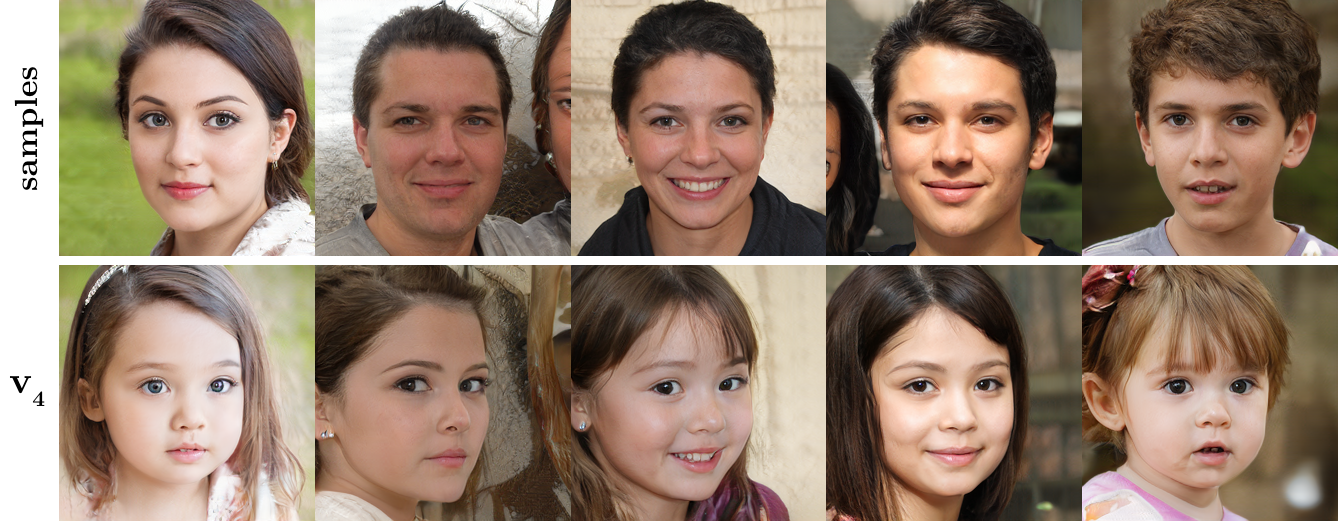}
    \caption{GANSpace editing is compatible with our overparameterized model. Latent translation of random generated images (row 1) along the fourth principal component (row 2), corresponding to an interesting combination of age, head rotation, and gender.} \label{fig:ganspace}
\end{figure}

\begin{figure*}[th!]
    \includegraphics[width=\linewidth]{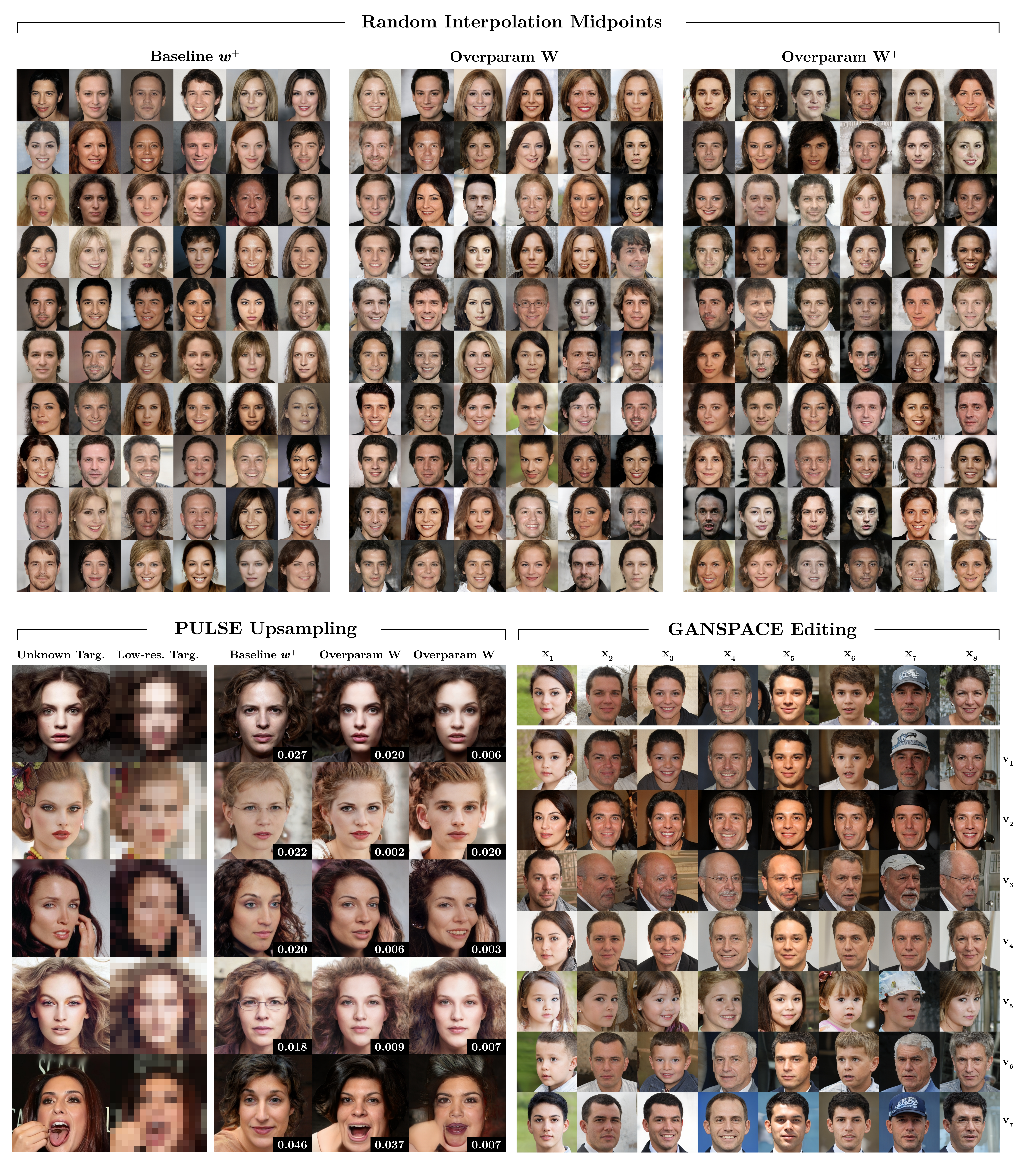}
    \caption{\textbf{Top Left}: Random interpolation midpoints for various methods, showing that 1) our approach is competitive with the baseline 2) extension to $\mathbf{W}^+$ damages interpolation quality. \textbf{Bottom Left}: Additional results for PULSE upsampling for random samples from the CelebA dataset, which also give full-resolution target. \textbf{Bottom Right}: Top 7 principal components for GANSpace in our latent space.}
    \label{fig:supplemental-editing}
\end{figure*}

Our method replaces a single vector $\boldsymbol{w} \in \mathcal{W}$ with 512 vectors $\mathbf{W}_i \in \mathcal{W}$. As described in sec.~\ref{method-overparam-modulation}, these vectors are correlated samples, but the $\mathcal{W}$ distribution remains unchanged. Hence, our approach is still compatible with unsupervised editing techniques such as GANSpace~\cite{ganspace} (fig.~\ref{fig:ganspace}), which extracts the principal components of $\mathcal{W}$. Please refer to fig.~\ref{fig:supplemental-editing} for more results.


\section{Limitations}

Overparameterizing the latent space to such an extent, while simple conceptually, can still be considered a considerable modification to the baseline model. This section investigates potential limitations associated with these changes.

\subsection{Computational efficiency}
\label{comp-eff}

\begin{table}[H]
\small
\begin{tabularx}{\linewidth}{Xcc}
\toprule
& \textbf{Memory}\ (Gb)$_\downarrow$& \textbf{Speed}\ (sec/kimg)$_\downarrow$\\
\midrule
Baseline & 3.90 & 8.33\\
Overparam & 4.90 & 8.80 \\
\bottomrule
\end{tabularx}
\caption{Memory usage and speed when training on four \textit{NVIDIA A100} cards, at resolution $256\times256$. Despite increased memory consumption, wall-clock time remains within 10\% of the baseline.
}
\label{table:performance}
\end{table}

Our revised style injection mechanism may seem inefficient as it requires 512 calls to the mapping and affine projection layers. It is true that memory consumption increases, by approximately 25\% during training, as shown in table \ref{table:performance}. However, actual run-time remains competitive despite an increased number of operations, which we attribute to cache-efficiency (these calls are easy to vectorize)\footnote{Note that in StyleGAN2 specifically, the baseline could be accelerated by training without fused modulation, which accelerates the PPL regularizer's gradient computation. As our approach is incompatible with unfused training, we did not train with this optimization. Besides, it is also omitted in StyleGAN3.}.

\subsection{Non-injectivity and non-determinism}

\begin{figure}[ht]
    \includegraphics[width=\linewidth]{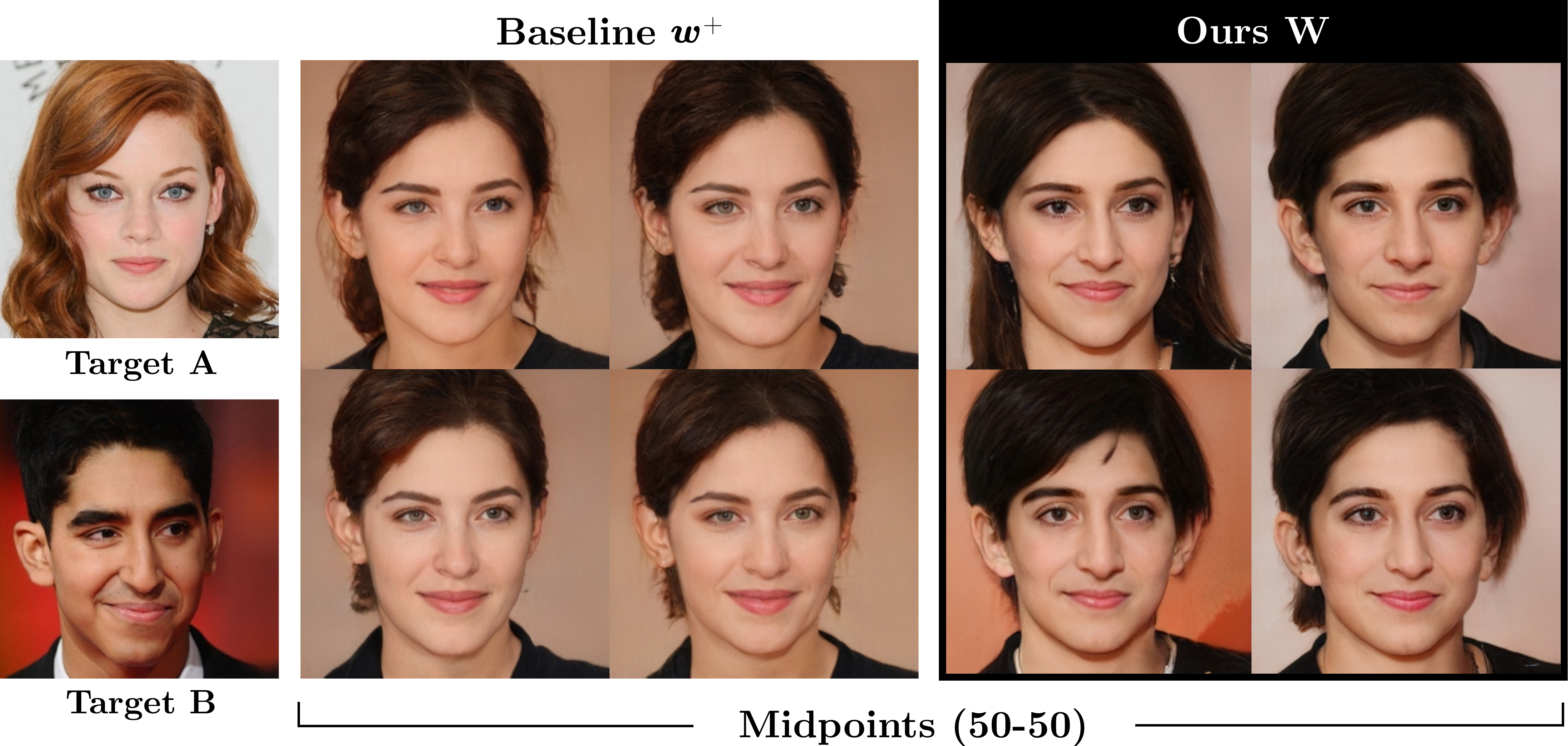}
    \caption{Overparameterization leads to a greater variety of latent codes when optimization is unregularized and initialized randomly. Interpolation between different codes for the same target images (left) in an overparameterized space yields more variable solutions (right) as compared to the baseline (middle). } \label{fig:nondet}
\end{figure}

StyleGAN is non-injective, meaning that multiple latent codes generate the same image~\cite{inversion-survey}. As it drastically increases the latent dimensionality, our approach exacerbates this effect. In fig.~\ref{fig:nondet}, we invert the same targets multiple times without regularization, after initializing to random points, and show that the interpolated images exhibit much more variance. The exact impact of this change on image editing using latent directions~\cite{interfacegan} remains to be explored.

\section{Discussion}

In the larger context of deep learning, it is well known that optimization is often surprisingly easy in high dimensional, non-linear spaces; local minima are not a problem in practice even when nothing guarantees their absence~\cite{qualitative, the-loss-surfaces}. One could argue that the success of the field was partly based on this empirical observation. While the exact underpinnings of this effect are still being clarified~\cite{local-minima-in-training, local-minima}, this observation is well-accepted and has shown to hold time and time again in practice. Perhaps even more surprising is the fact that these strongly overparameterized networks still generalize well~\cite{the-role-of-overparam, overparam-interpolation}. Remarkably, it has been shown that such networks do not overfit, provided that training is sufficiently long. This phenomenon was called "double descent"~\cite{double-descent} and goes against classical statistical intuition which would predict the opposite effect. In fact, the best-performing models are almost always grossly overparameterized~\cite{gpt-3, convnext, megatron}. 

While our work addresses overparameterizing the latent space of a network, rather than its parameter space, there are certainly interesting parallels to be drawn. To us, at least, it is  surprising that training is still possible following these changes, and that the obtained latent vectors are still useful. It is thought that in generative modeling, the generator's intrinsic dimensionality should ideally match that of the real image manifold~\cite{intro-dgn, improved-precision-recall}. While the latter is hard to calculate~\cite{on-noise-injection}, it has been estimated to values as low as 20--50~\cite{on-the-intrinsic-dimensionality, intrinsic-dimension}. Yet, training remains effective with our seemingly excessive overparameterization. Our work shows once again that intuition obtained from low-dimensional statistical models can be misleading. 

\section{Conclusion}
We propose overparameterization as an effective strategy for addressing StyleGAN inversion and show how it can be effectively implemented in practice. Our method trains efficiently and stably, and inverts well, finally achieving near-zero reconstruction error without compromising editability. The crux of our approach is that it avoids post-training extensions to the latent space, which trade-off between these two objectives. Finally, we demonstrate compatibility with existing downstream tasks. While our approach does not require encoders for great editability, and while it improves convergence speed, optimization is still too slow for real-time applications. We leave the design of an encoder architecture that targets our overparameterized latent space as future work.

\section*{Acknowledgements} {\small This research was supported by NSERC grants CRDPJ 537961-18 and RGPIN-2020-04799, and by Compute Canada.}

{
\small
\bibliographystyle{ieee_fullname}
\bibliography{refs}
}
\appendix
\onecolumn
\section{Supplementary Material}
\begin{figure}[h!]
    \centering
    \includegraphics[width=0.9\linewidth]{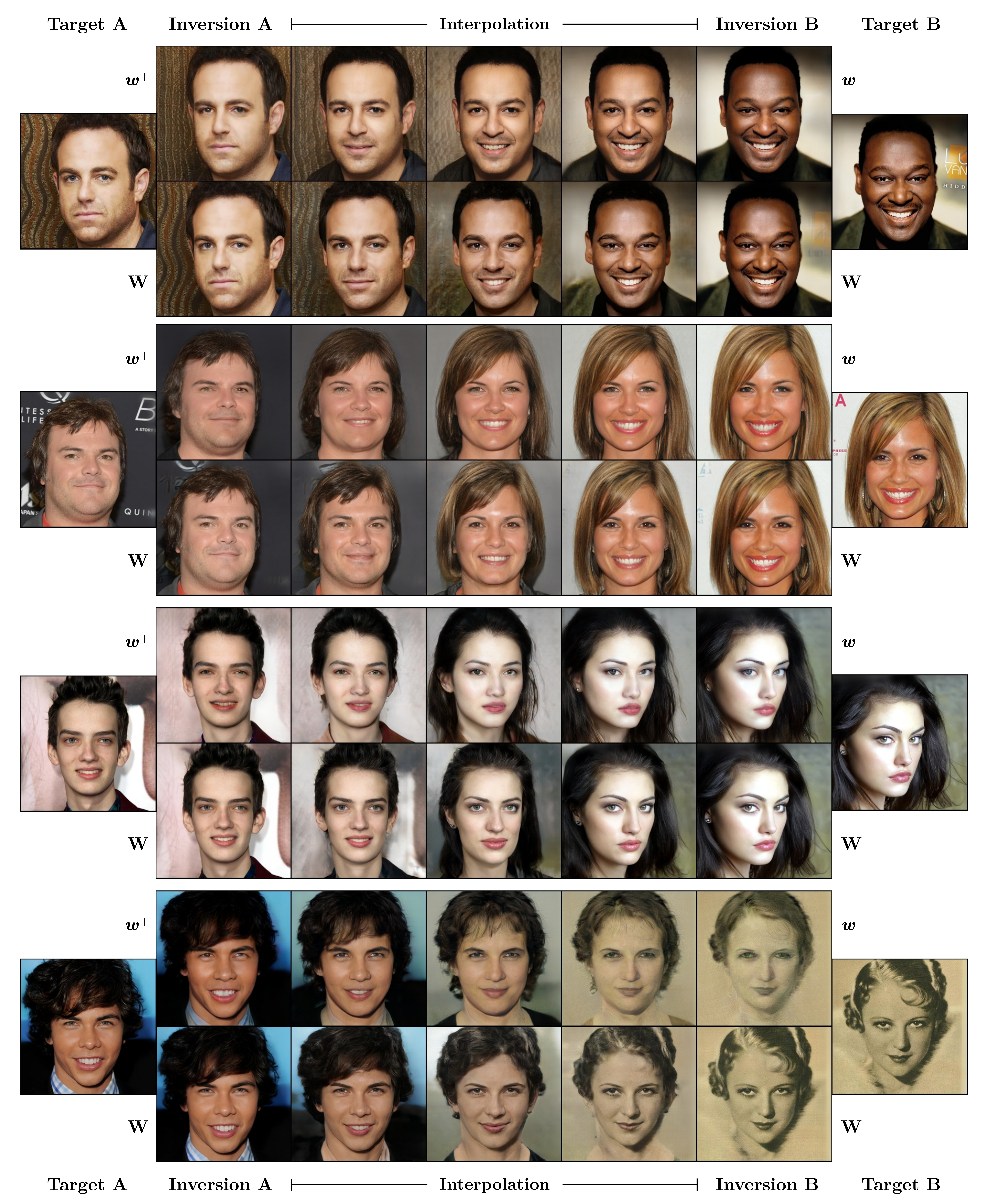}
    \caption{Additional interpolation results, comparing the baseline ($\boldsymbol{w}^+$, first row of each group) to our final approach ($\mathbf{W}$, second row of each group). Our method obtains much better reconstruction quality ("Inversion A/B" better match "Target A/B"), without compromising interpolation quality (the middle columns second row images are as realistic as the first row baseline). Note robustness to an out-of-sample image (vintage black and white portrait) in the last group. }
    \label{fig:supplemental-interpolation}
\end{figure}
\end{document}